\documentclass[a4paper]{article}

\usepackage{INTERSPEECH2022}
\usepackage{subcaption}
\usepackage{tipa}
\usepackage{float}
\usepackage{epstopdf}

\title{Cross-lingual Self-Supervised Speech Representations for \\ Improved Dysarthric Speech Recognition}

\name{Abner Hernandez$^{1}$, Paula Andrea P\'erez-Toro$^{1,2}$, Elmar N\"oth$^{1}$, Juan Rafael Orozco-Arroyave$^{1,2}$, Andreas Maier$^{1}$, Seung Hee Yang$^{3}$}

\address{Author Affiliation(s)}

\address{$^1$Pattern Recognition Lab. Friedrich-Alexander Universit{\"a}t Erlangen-N{\"u}rnberg, Germany\\ $^2$GITA Lab. Facultad de Ingenier\'ia. Universidad de Antioquia UdeA, Medell\'in, Colombia\\  
$^3$Speech \& Language Processing Lab. Friedrich-Alexander Universit{\"a}t Erlangen-N{\"u}rnberg, Germany} 

\email{abner.hernandez@fau.de, paula.andrea.perez@fau.de, elmar.noeth@fau.de, rafael.orozco@udea.edu.co, andreas.maier@fau.de, seung.hee.yang@fau.de}

\begin{document}

\maketitle
\begin{abstract}
State-of-the-art automatic speech recognition (ASR) systems perform well on healthy speech. However, the performance on impaired speech still remains an issue. The current study explores the usefulness of using Wav2Vec self-supervised speech representations as features for training an ASR system for dysarthric speech. Dysarthric speech recognition is particularly difficult as several aspects of speech such as articulation, prosody and phonation can be impaired. Specifically, we train an acoustic model with features extracted from Wav2Vec, Hubert, and the cross-lingual XLSR model. Results suggest that speech representations pretrained on large unlabelled data can improve word error rate (WER) performance. In particular, features from the multilingual model led to lower WERs than filterbanks (Fbank) or models trained on a single language. Improvements were observed in English speakers with cerebral palsy caused dysarthria (UASpeech corpus), Spanish speakers with Parkinsonian dysarthria (PC-GITA corpus) and Italian speakers with paralysis-based dysarthria (EasyCall corpus). Compared to using Fbank features, XLSR-based features reduced WERs by 6.8\%, 22.0\%, and 7.0\% for the UASpeech, PC-GITA, and EasyCall corpus, respectively.

\end{abstract}

\section{Introduction}

Individuals with Parkinson's, cerebral palsy, amyotrophic lateral
sclerosis (ALS), and other disorders greatly benefit from being able to use speech-enabled technology. However, most automatic speech recognition (ASR) systems are not able to accurately recognise speech from patients with dysarthric speech. Several reasons cause low dysarthric speech recognition performance such as imprecise articulation that deviates from standard healthy speech. Furthermore, dysarthric speech not only varies between speakers but can also vary within speakers and in severity. Dysarthria itself is subdivided into several types such as flaccid dysarthria, spastic dysarthria, ataxic dysarthria, hypokinetic dysarthria~\cite{darley1969}.

Another reason for degraded performance on dysarthric speech recognition is the scarcity of impaired speech data. While standard speech recognizers are trained on thousands of hours of speech data, dysarthric speech is more difficult to collect. A recent method of alleviating the issue of speech data scarcity is the use of self-supervised representation learning. This method involves pretraining a network on a large unlabelled corpus. The network learns speech representations from raw audio which can be used in downstream tasks such as speech recognition. Speech representation learning models such as 
Wav2Vec2.0~\cite{Baevski2020}, and Hubert~\cite{Hsu2021Hubert}, 
have shown that learned representations produce state-of-the-art results on a variety of speech tasks: Speaker and language identification~\cite{fan21_interspeech}, emotion recognition~\cite{pepino21}, spoofing speech detection~\cite{xie21_interspeech}, and second language mispronunciation detection~\cite{xu21k_interspeech}. However, there is a lack of studies regarding the benefits of Wav2Vec or other speech representation models for impaired speech tasks like speech recognition or diagnosis.

\subsection{Self-Supervised Representation Learning}
\begin{figure}[htb]
    \centering
    \centerline{\includegraphics[width=7.0cm]{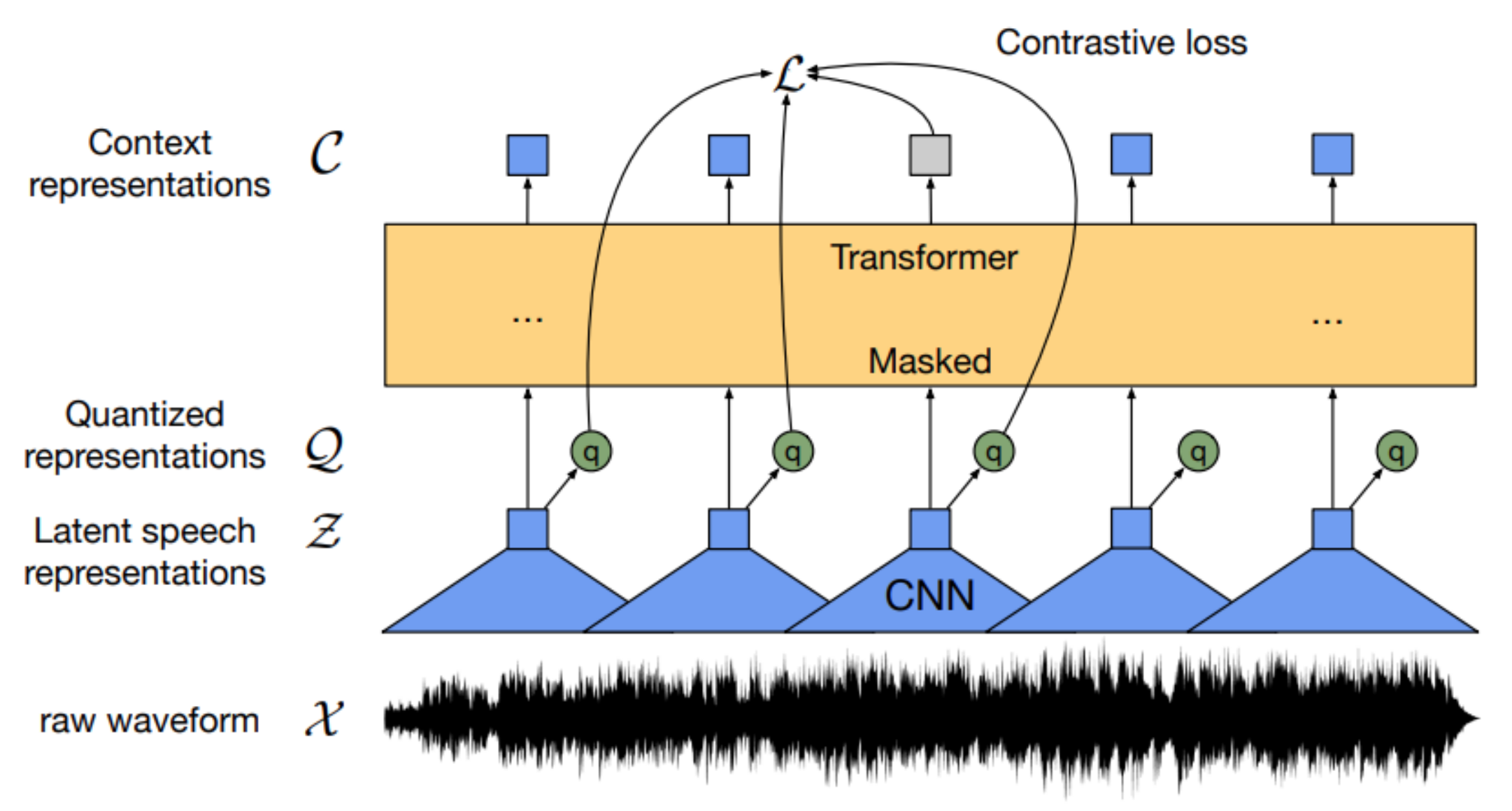}}
    \caption{Wav2Vec 2.0 framework for learning speech representations~\cite{Baevski2020}.}
    \label{fig:wav2vec2}
\end{figure}

Since labelled data are difficult to obtain, research into pre-training with unlabelled data has grown. In general, self-supervised representation learning involves training a model on a large amount of unlabelled data and learning an acoustic representation to be used on a downstream task. For example, Figure~\ref{fig:wav2vec2} displays the Wav2Vec2.0 model which takes a raw waveform as input to several blocks of temporal convolutions. The output is then masked and fed to a transformer-based context network that uses convolutional layers to learn relative positional embeddings. The output of this feature encoder utilizes product quantization to select quantized representations which can then be used on a downstream task in replacement of traditional features such as Mel-frequency cepstral coefficients (MFCC). The training process depends on a contrastive loss where the model needs to identify the true quantized speech representation. Other approaches include using BERT-like masked prediction (Hubert)~\cite{Hsu2021Hubert} or masked reconstruction (TERA)~\cite{liu2021tera}.

\subsection{Related Work}
Several techniques have been proposed to handle the lack of dysarthric speech data. Augmenting healthy speech data to replicate dysarthric speech was explored in~\cite{vachhani2018,jiao2018}. In~\cite{xiong2019}, a speaker-dependent phonetic analysis-based augmentation method was proposed. A generative adversarial network(GAN) based approach that directly modifies dysarthric speech to sound more healthy showed Word Error Rate (WER) improvements in~\cite{biosignals20}. 

WER is not only useful for assessing speech recognition systems but also for assessing speech impairments~\cite{kitzing2009automatic}. The PEAKS system for evaluating speech and voice disorders relies on automatic speech recognition for analyzing impairments in laryngectomees and children with cleft lip and palate~\cite{maier2009peaks}.
In~\cite{vasquez2016word}, two features based on the word accuracy and the dynamic time warping algorithm were proposed to assess intelligibility deficits in Parkinson's disease (PD) patients.
The suitability of these features was evaluated on the PC-GITA corpus~\cite{orozco2014new} for the recognition of words in several sentences and used to discriminate between PD patients and healthy control (HC) subjects (accuracies of up to 92\%).
Similarly, an ASR system based on Hidden Markov Models (HMM) was proposed in~\cite{parra2018automatic}. The authors evaluated this approach in the same corpus but in diadochokinetic tasks, where WERs of 2.90 for HC subjects and 7.10 for PD patients were reported.

Pretrained speech representations for impaired speech is a relatively unexplored field. Positive results were obtained for Alzheimer's detection through speech with Wav2Vec~\cite{zhu21,gauder21}. However, limited studies have evaluated the effectiveness of speech representations for dysarthric speech recognition. The current study explores using Wav2Vec speech representations to handle both the variability of dysarthric speech and the scarceness of data. In particular, the Wav2Vec 2.0 model trained on cross-lingual data (XLSR) should have been exposed to a large variety of similar phonemes that could be useful for dysarthric speech recognition.
The rest of the paper is organized as follows. Section 2 presents the Spanish and English dysarthric speech datasets. The experimental procedures and results are presented in Sections 3 and 4. Lastly, Section 5 concludes the paper with some future directions. 

\section{Dysarthric Speech Databases}
\label{sec:data}

\subsection{UASpeech}
The UA-Speech corpus~\cite{kim08} is composed of 15 patients with cerebral palsy and 13 healthy controls. Participants were recruited by the Rehabilitation Education Center at
the University of Illinois between the ages of 19 and 58. The data is split into three blocks. Each block contains 255 words from all 28 speakers. From those 255 words, the same 10 digits, 26 radio alphabet words, 19 computer commands and 100 common words occur in each block. There are also 300 uncommon words that are split between the three block such that each block contains different uncommon words.

Speech intelligibility was assessed to obtain the severity of dysarthria for each speaker. Five naive listeners were instructed to provide orthographic transcriptions of each word spoken by an individual with dysarthria. The correct percentage was then averaged across five listeners to obtain each speaker’s intelligibility. Speakers were classified into very low (0-25\%), low (26\%-50\%), mid (51\%-75\%) and high (76\%-100\%) levels of intelligibility.

\subsection{Extended Version of PC-GITA}

This dataset is part of an extended version of the PC-GITA corpus~\cite{orozco2014new}. 
It considers speech recordings from 50 PD patients and 50 HC subjects in the same acoustic conditions, and 20 PD and 20 HC in different non-controlled acoustic conditions. All participants are native Colombian Spanish speakers that were asked to read a total of 21 isolated words.
The patients were evaluated by an expert neurologist according to the third part of the Movement Disorders Society Unified Parkinson's Disease Rating Scale~(MDS-UPDRS-III) which ranges from 0 to 152, where 0 means completely healthy and 152 means highly impaired~\cite{Goetz2008}. All participants were evaluated by three different phoniatricians according to the modified Frenchay Dysarthria Assessment~(mFDA)~\cite{Vasquez18-TAA} scale, which
ranges from 0 to 52, where 0 means healthy speech and 52 means
highly impaired speech production. For the labelling of the
participants the median between the three phoniatricians was 
considered.
MDS-UPDRS-III aims to assess the neurological state of the PD patients, while mFDA evaluates the dysarthria level.
Additional demographic information of the participants is presented in Table~\ref{tab:pc-gita}.

\begin{table}[!htpb]
\centering
\caption{General demographic information of the subjects in the extended PC-GITA dataset. Higher UPDRS and FDA scores reflect more impairments in speech.} 

\label{tab:pc-gita}
\resizebox{0.83\linewidth}{!}{
\begin{tabular}{lcc}

\multicolumn{1}{c}{\textbf{}}  & PD patients         & HC subjects \\ 
\multicolumn{1}{c}{\textbf{}}  & F\,/\,M         & F\,/\,M \\ \toprule
Number of subjects               & 36\,/\,34                & 34\,/\,36                             \\
Age [years]                  & 59.8\,/\,62.6  & 59.4\,/\,62.8                 \\
MDS-UPDRS-III     & 36.4\,\,/\,40.4\, &    --                      \\
mFDA      & 28.1\,/\,29.3 &   6.6\,/\,8.7                                      \\ \bottomrule
\multicolumn{3}{l}{Values are expressed as mean. F: female. M: male.}\\
\multicolumn{3}{l}{The MDS-UPDRS-III ranges from 0 to 132 and}\\
\multicolumn{3}{l}{the mFDA ranges from 0 to 52. } \\
\end{tabular}
}
\end{table}

\subsection{EasyCall}
The EasyCall corpus consists of 24 healthy (10 females, 14 males) and 31 dysarthric (11 females, 20 males) Italian speakers~\cite{turrisi21}. A range of disorders causing dysarthria includes Parkinson’s Disease, Huntington's Disease, Amyotrophic Lateral Sclerosis, peripheral neuropathy, myopathic or myasthenic lesions. However, among the 31 patients, 24 speakers have paresis (partial paralysis) based dysarthria. The severity of the dysarthria was assessed with the Therapy Outcome Measure (TOM)~\cite{enderby2013} by a neurologist. The TOM score ranges from 1 to 5 and corresponds to mild, mild-moderate, moderate, moderate-sever, and severe
dysarthria.
The corpus is specifically created to improve smartphone ASR systems and includes simple mobile command phrases like "scroll up" (Scorri verso l’alto), "new contact" (Nuovo contatto) , "end call" (Termina chiamata). Speakers recorded 66 to 69 sentences using an in-house smartphone application. 

\section{Experiment}
\label{sec:majhead}
The training process with the UA-Speech data follows from previous studies using the database~\cite{xiong2019}. All audio from the 13 healthy controls is used for training, along with block 1 and block 3 from dysarthric speakers. Testing is conducted on block 2 of dysarthric speakers. We also train a model solely on control speakers. Unlike previous studies using UA-Speech, we do not trim silences from the audio.

With the PC-GITA corpus, we run two main experiments. First, we train with all control speakers and then test with dysarthric speakers from the same recording session. We also evaluate the model with dysarthric speakers from a second recording session and different acoustic conditions. Second, we train an acoustic model with both control and dysarthric speakers from recording session 1 and evaluate on different dysarthric speakers from the second recording session.

Similar to the PC-GITA corpus, the EasyCall experiments include a training session with only control speakers and another session with control and dysarthric speakers (no speaker in the test set in the train set). The train and test split was selected based on the split from the EasyCall article~\cite{turrisi21}. The test set contains 12 dysarthric speakers (7 male, 5 female) ranging from mild to severe dysarthria. 

Speech features are extracted using the base and large Wav2Vec2.0 model~\cite{Baevski2020}, the multilingual XLSR model~\cite{conneau21}, and Hubert~\cite{Hsu2021Hubert}. Large Wav2Vec and Hubert models are pretrained on 60,000 hours of English data from the LibriVox database. The XLSR model is trained on 56,000 hours of audio from 53 different languages using the Multilingual Librispeech, CommonVoice, and Babel data sets. The UA-Speech corpus is trained with Wav2Vec, Hubert and XLSR models, while the PC-GITA is only trained with the XLSR model. 
Acoustic models are built using the ESPnet~\cite{watanabe2018} toolkit. All acoustic models are trained end-to-end with a conformer encoder and transformer decoder~\cite{gulati20}. We use Byte Pair Encoding sub-words as output units. 500 sub-words for English, and 178 for Spanish. No language model is used when decoding. All experiments use the same acoustic model with only changes in learning rate. 

\section{Results}
\label{sec:Results}
\subsection{UA-Speech Results}
Results in Table~\ref{tab:tab2} show that all self-supervised representation trained models outperform baseline filterbank (Fbank) features. The Wav2Vec 2.0 model showed WER improvements of 29.3\% compared to the Fbank model's 32.9\%. However, the best performing model is the multilingual XLSR model with a WER of 26.1\% and a character error rate (CER) of 24.1\%. Results from individual intelligibility categories are examined in Table~\ref{tab:tab3}. These results reflect the speaker-dependent models from Table~\ref{tab:tab2}. Improvements were seen across all categories. Compared to Fbanks, XLSR feature reduce WER by 5.4\% for very low, 8.9\% for low, 12.2\% for mid, and 3.8\% for high intelligibility speakers.


\begin{table}[th]
\caption{Performance of speech representation with the UA-Speech data when conducting speaker-dependent training.}
\label{tab:tab2}
\centering
\resizebox{0.70\linewidth}{!}{
\begin{tabular}{c c c}

\multicolumn{1}{c}{Model} & 
\multicolumn{1}{c}{WER(\%)} &
\multicolumn{1}{c}{CER(\%)}\\
\toprule
Fbank & 32.9 & 30.8 \\
Wav2Vec 2.0 & 29.3 & 27.7 \\
Hubert & 29.7 & 28.2 \\
XLSR  & \textbf{26.1} & \textbf{24.1} \\
\bottomrule
\end{tabular}
}
\end{table}

\begin{table}[th]
\caption{UA-Speech speaker-dependent WERs (\%) of speech representation based on intelligibility categories.}
\label{tab:tab3}
\centering
\resizebox{0.75\linewidth}{!}{
\begin{tabular}{c c c c c}

\multicolumn{1}{c}{Model} & 
\multicolumn{1}{c}{Very low} &
\multicolumn{1}{c}{Low} &
\multicolumn{1}{c}{Mid} &
\multicolumn{1}{c}{High}\\

\toprule\
Fbank & 67.4 & 37.4 & 31.5 & 9.9 \\
Wav2Vec 2.0 & 66.7 & 33.3 & 36.3 & 6.7 \\
Hubert & 68.0 & 36.0 & 23.4 & \textbf{6.2} \\
XLSR  & \textbf{62.0} & \textbf{28.6} & \textbf{19.3} & 6.2 \\
\bottomrule
\end{tabular}
}
\end{table}

According to Table~\ref{tab:tab4}, XLSR extracted features are the most robust for speaker-independent models only trained on healthy speech. Compared to the 63.9\% WER from using Fbanks, Hubert and XLSR based features achieved WERs of 50.3\% and 47.3\% respectively. In general, features extracted from the multilingual XLSR model outperformed Wav2Vec, Hubert, and Fbank features for all English experiments. 

\begin{table}[th]
\caption{Speaker-independent WER performance of speech representation with the UA-Speech data.}
\label{tab:tab4}
\centering
\resizebox{0.70\linewidth}{!}{
\begin{tabular}{c c c c}

\multicolumn{1}{c}{Model} & 
\multicolumn{1}{c}{WER(\%)} &
\multicolumn{1}{c}{CER(\%)}\\

\toprule\
Fbank & 63.9 & 64.4 \\
Wav2Vec 2.0 & 47.6 & \textbf{44.3} \\
Hubert & 50.3 & 47.4 \\
XLSR  & \textbf{47.3} & 45.0 \\
\bottomrule
\end{tabular}
}
\end{table}

\subsection{PC-GITA Results}
Results with the PC-GITA data is limited to the multilingual XLSR model. Training sets for all conditions use healthy speech data from recording session 1, while testing is conducted on either session 1 or session 2 data from dysarthric speakers. XLSR-based features show improvements in both session 1 and 2 compared to Fbanks, see Table~\ref{tab:pcgita-results}. The effectiveness of XLSR features is further emphasized by the superior performance on session 2 which contains recordings with different speakers and different acoustic conditions. Further improvements are seen by XLSR-PD when including speech from PD patients. A CER of 4.5\% was achieved when training on healthy speech and PD speech from session 1. Table~\ref{tab:pcgita-results2} displays results when dividing speakers by MDS-UPDRS III and mFDA scores. As there is no official or standard method of classifying speakers into severity levels we simply divide speakers into three groups (low, mid, and high). Depending on the measure type, CERs range 8.5\% to 25.5\% when training with Fbanks and between 1.7\% and 10.9\% when training on XLSR features. 

\begin{table}[th]
\caption{Fbank and XLSR models are trained on healthy speech data only. XLSR-PD, and Fbank-PD are trained on both healthy and dysarthric speech from Session one.}
\label{tab:pcgita-results}
\centering
\begin{tabular}{ccccc}
Model                                                      & Data  & WER(\%)                                             & CER(\%)                                             & Test Sess. \\ \toprule
\begin{tabular}[c]{@{}c@{}}Fbank\\ XLSR\end{tabular}       & 58min & \begin{tabular}[c]{@{}c@{}}20.9\\ 6.10\end{tabular} & \begin{tabular}[c]{@{}c@{}}13.7\\ 4.3\end{tabular}  & 1          \\ \hline
\begin{tabular}[c]{@{}c@{}}Fbank\\ XLSR\end{tabular}       & 58min & \begin{tabular}[c]{@{}c@{}}46.6\\ 28.1\end{tabular} & \begin{tabular}[c]{@{}c@{}}32.5\\ 16.1\end{tabular} & 2          \\ \hline
\begin{tabular}[c]{@{}c@{}}Fbank-PD\\ XLSR-PD\end{tabular} & 70min & \begin{tabular}[c]{@{}c@{}}34.9\\ 12.9\end{tabular} & \begin{tabular}[c]{@{}c@{}}20.8\\ 4.5\end{tabular}  & 2          \\ \bottomrule
\end{tabular}
\end{table}

\begin{table}[th]
\caption{Severity-based performance (CER\%). UPDRS scores were divided into low (0-30), mid (31-60), and high (61-92). Similarly, mFDA scores are divided as low (0-17), mid (18-34), and high (35-52). Higher scores reflect more impairments in speech.}
\label{tab:pcgita-results2}
\centering
\centering
\begin{tabular}{ccccc}
Model & Scale & Low & Mid & High\\ 
\toprule
Fbank & \begin{tabular}[c]{@{}c@{}}MDS-UPDRS-III\\ mFDA\end{tabular} & \begin{tabular}[c]{@{}c@{}}8.5\\ 9.8\end{tabular} & \begin{tabular}[c]{@{}c@{}}15.1\\ 10.4\end{tabular} & \begin{tabular}[c]{@{}c@{}}25.5\\ 23.8\end{tabular} \\ \hline
XLSR  & \begin{tabular}[c]{@{}c@{}}MDS-UPDRS-III\\ mFDA\end{tabular} & \begin{tabular}[c]{@{}c@{}}1.7\\ 4.1\end{tabular} & \begin{tabular}[c]{@{}c@{}}4.8\\ 2.7\end{tabular}   & \begin{tabular}[c]{@{}c@{}}10.9\\ 8.0\end{tabular}  \\ \bottomrule
\end{tabular}
\end{table}

\subsection{EasyCall Results}
\vspace{-0.1cm}
Similar to the PC-GITA results, acoustic models trained on XLSR-based features outperform traditional Fbank features presented in Table~\ref{tab:easycall-results}. Furthermore, results from our experiments show lower WER's than the commercial ASR systems (Microsoft and IBM speech-to-text) used for evaluation in the EasyCall article. In the case of training on healthy speech only, XLSR trained systems reduced the WER from 31.8\% to 24.8\%, while including dysarthric speech data further decreases the WER to 16.5\%. 

\begin{table}[th]
\caption{XLSR-PD, and Fbank-PD are trained on both healthy and dysarthric speech from Session one.}
\label{tab:easycall-results}
\centering
\begin{tabular}{c c c c}
Model & Train Data  & WER(\%) & CER(\%)\\ 
\toprule
EasyCall ASR1~\cite{turrisi21} & - & 61.90 & -\\
EasyCall ASR2~\cite{turrisi21} & - & 135.26 & -\\
\hline
Fbank & 4 hours  & 31.8 & 25.4\\
XLSR &  4 hours & \textbf{24.8} & \textbf{19.6}\\
\hline
Fbank-PD & 9 hours & 22.6 & 17.3\\
XLSR-PD & 9 hours  & \textbf{16.5} & \textbf{12.1}\\
\bottomrule
\end{tabular}
\end{table}

\begin{table}[th]
\caption{EasyCall CERs(\%) based on TOM score ranges. Results are from models trained only with healthy speech}
\label{tab:easycall-results2}
\centering
\begin{tabular}{c c c c}
Model & Low & Mid & High\\
\toprule
Fbank & 11.8 & 31 & 88.8\\
XLSR  & 7.2 & 28.4 & 73.9\\
\bottomrule
\end{tabular}
\end{table}

\subsection{Visualization}
\begin{figure}[th]
  \centering
  \begin{subfigure}{.25\textwidth}
    \includegraphics[width=\columnwidth]{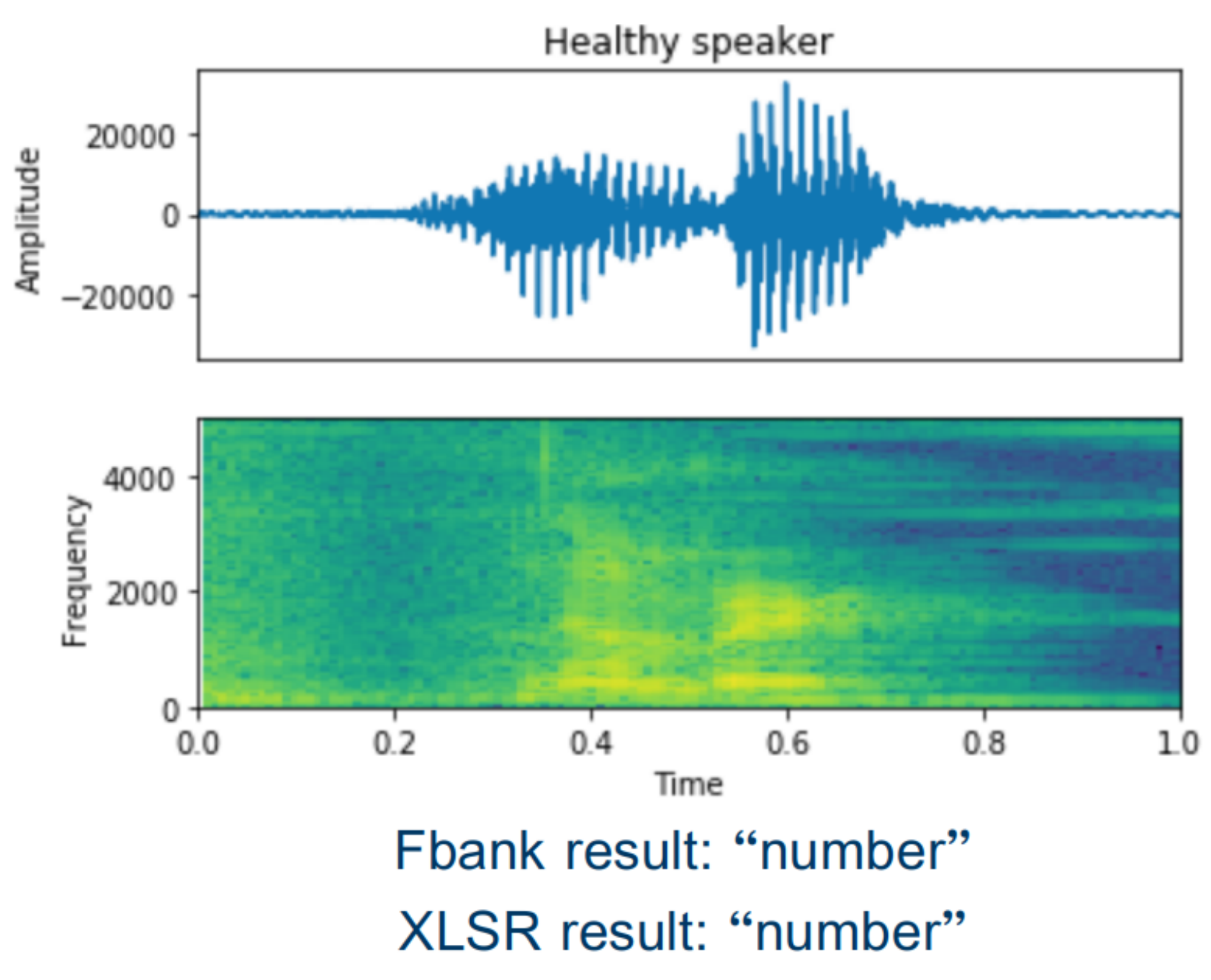}
  \end{subfigure}
  \begin{subfigure}{.25\textwidth}
    \includegraphics[width=\columnwidth]{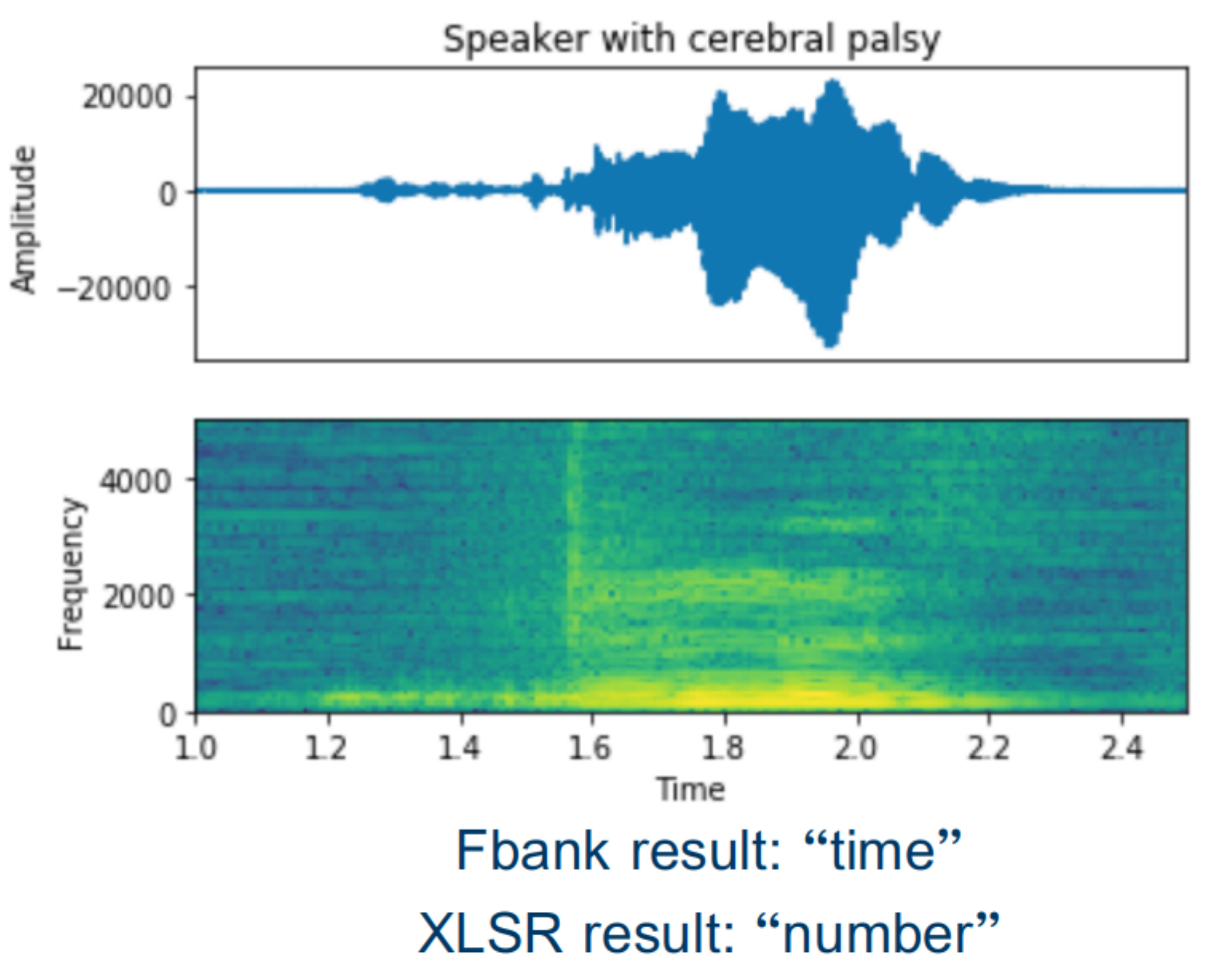}
  \end{subfigure}
  \caption{Spectrogram of the word "number" for a healthy (top) and dysarthric (bottom) speaker.}
\label{fig:spec}
\end{figure}

Figure~\ref{fig:spec} shows the spectrogram of a healthy speaker and a speaker with dysarthria producing the word "number". From the healthy speaker's waveform and spectrogram, we see a clear distinction between the two syllables in \textipa{/n2m.b@r/}. Both Fbank, and XSLR trained models correctly predict the correct word number. The distinction between the two syllables in the word number is less apparent in the speaker with dysarthria. Instead, the two syllables are blended together and an Fbank trained model incorrectly predicts the word as "time". While this prediction is very wrong, by looking at the spectrogram it is understandable prediction. The first phoneme in \textipa{/taIm/} is the stop consonant /t/ which is often visualized as a sudden burst. This is followed by a diphthong vowel which tends to be represented as a long segment.

\begin{figure}[th]
    \centering
    \centerline{\includegraphics[width=5cm]{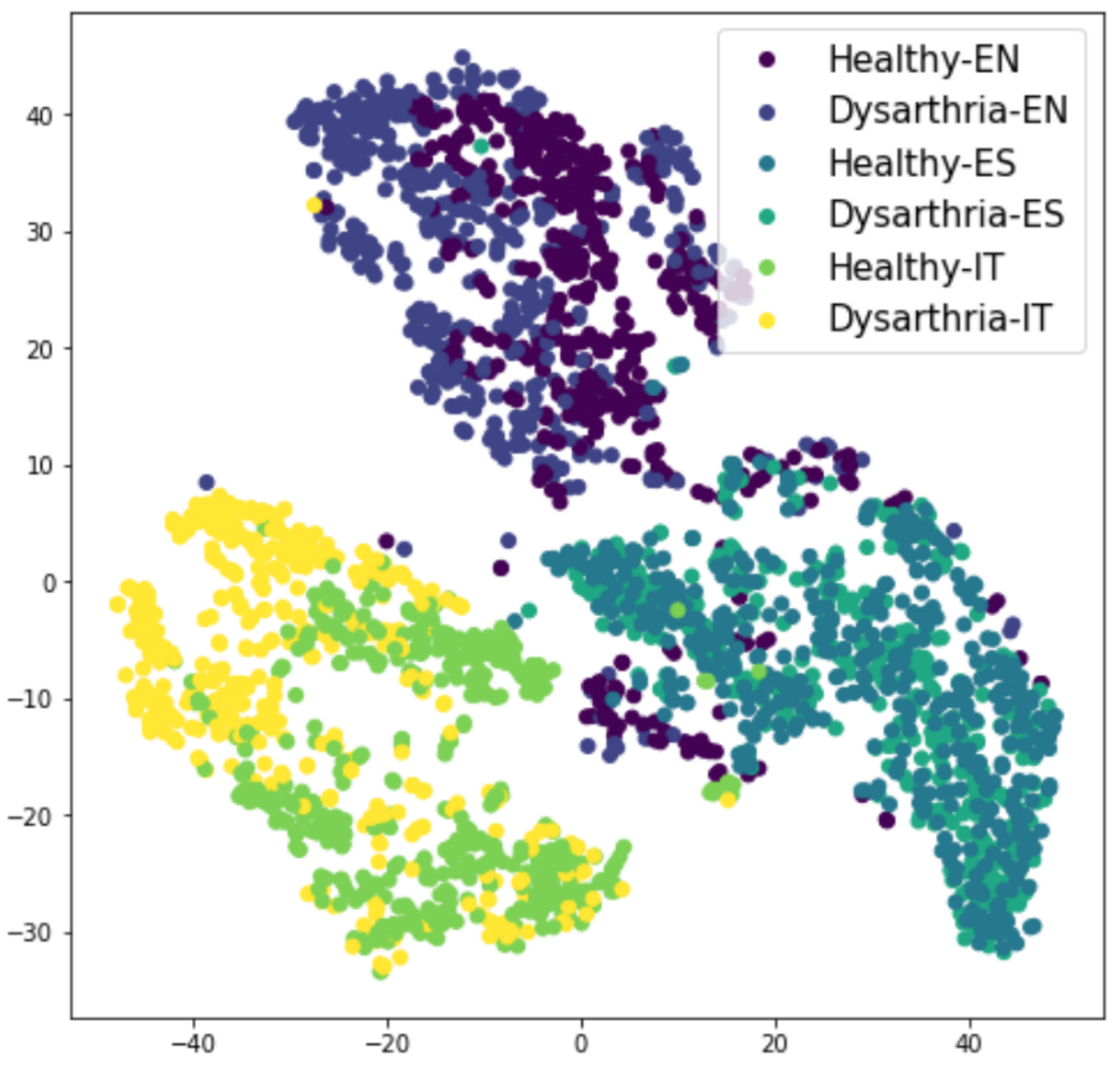}}
    \caption{English, Spanish and Italian T-SNE~\cite{van2008visualizing} embedding representation of isolated words using XLSR raw embeddings.}
    \label{fig:raw-emb}
\end{figure}
Second, we can examine the raw XLSR embeddings used to train the acoustic model. From Figure~\ref{fig:raw-emb}, we see that the multilingual model clusters the three languages separately. Furthermore, while healthy and dysarthric speakers from PC-GITA show a large overlap in their representations, English and Italian speakers display slight non-overlapping clusters, possibly due to more severe dysarthria in the UASpeech and EasyCall.



\section{Conclusion}
\label{sec:page}
This paper evaluates the effectiveness of self-supervised representation learning for dysarthric speech recognition. We train acoustic models with features extracted from Wav2Vec2.0, Hubert, and XLSR models. Features extracted from the multilingual XLSR model produced the lowest WERs for all English, Spanish and Italian data sets. Despite that Wav2Vec and Hubert models were pretrained on 60,000 hours of English, XLSR features still led to lower WERs. Given that a multilingual model should contain more variations of similar phonemes, it may be more suitable for dysarthric speech recognition which can also vary. Results with the UA-Speech data show state-of-the-art performance, outperforming recent studies using the same data~\cite{xiong2019,vachhani2018,liu20}. Promising results were also obtained with the PC-GITA data. Particularly, good generalization with data from PD speakers recorded in different recording conditions. Also, fewer errors when training on less than an hour of healthy speech. Future work would benefit from exploring Wav2Vec representations more thoroughly and investigating whether useful information can be extracted for clinical therapy. Early detection of degenerative diseases using speech representations seems to be a promising direction for future research.

\bibliographystyle{IEEEtran}
\bibliography{mybib}

\end{document}